\documentclass[lettersize,journal]{IEEEtran}
\usepackage{amsmath,amsfonts}
\usepackage{algorithmic}
\usepackage{algorithm}
\usepackage{array}
\usepackage[caption=false,font=normalsize,labelfont=sf,textfont=sf]{subfig}
\usepackage{textcomp}
\usepackage{stfloats}
\usepackage{url}
\usepackage{verbatim}
\usepackage{graphicx}
\usepackage{cite}
\usepackage{hyperref}
\usepackage{adjustbox}
\usepackage{mathrsfs}
\usepackage{booktabs}
\usepackage{multirow}
\begin{document}

\title{Investigating Large Language Models for Code Vulnerability Detection: An Experimental Study}

\author{Xuefeng Jiang*, Lvhua Wu*, Sheng Sun, Jia Li, Jingjing Xue, Yuwei Wang, Tingting Wu, Min Liu$\dagger$
\thanks{Xuefeng Jiang, Lvhua Wu and Jingjing Xue are with the Institute of Computing Technology, Chinese Academy of Sciences and the University of Chinese Academy of Sciences (e-mail: \href{mailto:jiangxuefeng21b@ict.ac.cn}{jiangxuefeng21b@ict.ac.cn}) and \href{mailto:wulvhua24s@ict.ac.cn}{wulvhua24s@ict.ac.cn}.}
\thanks{Sheng Sun and Yuwei Wang are with the Institute of Computing Technology, Chinese Academy of Sciences (e-mail: \href{mailto:liumin@ict.ac.cn}{sunsheng@ict.ac.cn} and \href{mailto:ywwang@ict.ac.cn}{ywwang@ict.ac.cn}).}
\thanks{Jia Li is with the JD.com, Inc. (e-mail: \href{mailto:lijia1999@iie.ac.cn}{lijia1999@iie.ac.cn}).}
\thanks{Tingting Wu is with the China Mobile Research Institute. (e-mail: \href{mailto:wutingtingyjy@chinamobile.com}{wutingtingyjy@chinamobile.com}).}
\thanks{Min Liu is with the Institute of Computing Technology, Chinese Academy of Sciences, the University of Chinese Academy of Sciences and the Zhongguancun Lab (e-mail: \href{mailto:liumin@ict.ac.cn}{liumin@ict.ac.cn}).}
\thanks{* Xuefeng Jiang and Lvhua Wu share equal contributions to this work (sorted by author surname).}
\thanks{$\dagger$ Corresponding author: Min Liu}
}


\markboth{Journal of \LaTeX\ Class Files,~Vol.~14, No.~8, August~2021}%
{Shell \MakeLowercase{\textit{et al.}}: A Sample Article Using IEEEtran.cls for IEEE Journals}


\maketitle

\begin{abstract}
Code vulnerability detection (CVD) is essential for addressing and preventing system security issues, playing a crucial role in ensuring software security.
Previous learning-based vulnerability detection methods rely on either fine-tuning medium-size sequence models or training smaller neural networks from scratch. 
Recent advancements in large pre-trained language models (LLMs) have showcased remarkable capabilities in various code intelligence tasks including code understanding and generation. 
However, the effectiveness of LLMs in detecting code vulnerabilities is largely under-explored.
This work aims to investigate the gap by fine-tuning LLMs for the CVD task, involving four widely-used open-source LLMs. 
We also implement other five previous graph-based or medium-size sequence models for comparison. 
Experiments are conducted on five commonly-used CVD datasets, including both the part of short samples and long samples.
In addition, we conduct quantitative experiments to investigate the class imbalance issue and the model's performance on samples of different lengths, which are rarely studied in previous works.
To better facilitate communities, we open-source all codes and resources of this study in \href{https://github.com/SakiRinn/LLM4CVD}{https://github.com/SakiRinn/LLM4CVD} and \href{https://huggingface.co/datasets/xuefen/VulResource}{https://huggingface.co/datasets/xuefen/VulResource}.
\end{abstract}


\section{Introduction}
Detecting vulnerabilities in source codes is essential in protecting software applications from potential security risks. 
With the increasing number of vulnerabilities within today’s software, automating the detection process is becoming more and more critical for organizations to quickly respond and mitigate potential risks \cite{spw}.
Traditional methods mainly analyze the code vulnerability existence by dynamically executing the code program and observing the program output, with the assistance of fuzzing and symbolic execution techniques. 
In recent years, deep learning based static code vulnerability detection approach becomes one prominent research direction in security related communities. 
This approach often solely analyzes the code content, and does not require the execution of the code, which lowers the overhead to identify whether the code is vulnerable.
Early attempts include training graph-based models or sequence-based models. 

The graph-based models, represented by Devign \cite{devign}, attempt to transform the source code into the code graph, extract the code elements as graph nodes, and analyze vulnerabilities through graph representation learning. The sequence-based models, represented by CodeBERT \cite{codebert}, aim to regard the source code as a sequence of tokens, and utilize RNN-based or more advanced Transformer-based pre-trained language models to capture the vulnerable pattern within the code. The graph-based models are good at capturing the structural information of the code but struggle to capture long-distance association among the nodes, especially when the code content gets larger. 
Meanwhile, recent studies \cite{longtailed,xinzhou} and our fine-grained statistics across five commonly-used datasets in Table  \ref{tab:datasets} point out that the vulnerable code pattern tends to exist in the long code context. 
Thus, more efforts are put to the sequence-based models to detect code vulnerabilities, especially the pre-trained language models \cite{codebert}.

Large pre-trained language models (LLMs), as more powerful pre-trained language models, get remarkable successes in many general  downstream tasks like machine translation \cite{bayling} or code generation \cite{codeIntelligence}. 
However, few works explore whether the LLMs are capable to identify the code vulnerability, especially the fine-tuned LLMs on the CVD datasets \cite{xinzhou}.
For the code vulnerability detection (CVD) task, related representative works \cite{xinzhou,nong,zhangchao} aim to fix the model weights and design specific prompts to evaluate the performance on the close-sourced LLMs like ChatGPT and the open-sourced LLMs like the Llama series \cite{Llama-2}. One recent work VulLLM firstly tries to fine-tune the open-source LLMs but misses to incorporate evaluation on the longer code samples ($\textgreater$512 tokens), where vulnerable code patterns tend to exist  as referred in \cite{stagedvulbert}. 
In the meantime, experimental datasets are not unified in previous related works \cite{xinzhou,zhangchao,nong,vulllm}.

In this work, to bridge the above existing gap, we provide an early experimental investigation on fine-tuning LLMs on the CVD datasets, particularly focusing on 4 widely-used open-source Llama-series models, including two rarely evaluated LLMs (i.e. Llama-3 and Llama-3.1 \cite{Llama-3}). 
We revisit related literature, choose 5 most commonly-used CVD datasets, and additionally integrate 3 graph-based models and 2 medium-size BERT based sequence models into a unified codebase. We also study the impacts of class imbalance and code sequence length to the model performance with quantitative experiments.
In addition, all source codes with clear hand-on guidance are already open-source to facilitate related communities for more convenient reproduction of corresponding models.

To sum up, our contributions can be summarized as follows:
\begin{itemize}
\item We conduct a systematic investigation into the capabilities of fine-tuned LLMs for code vulnerability detection. Through comprehensive experiments and analysis, we evaluate the performance of 4 LLMs across 5 distinct code vulnerability datasets, involving the largest number of datasets among existing empirical studies. Furthermore, we compare their effectiveness with 3 representative graph-based model and 2 medium-size pre-trained sequence models. 
\item  We focus on the impact of datasets and hyperparameters on using LLMs for code vulnerability detection, both of which have often been neglected in prior research. We quantitatively demonstrate the impact of the positive sample ratio and sample length on fine-tuning LLMs by meticulously designed dataset resampling, as well as conduct a sensitivity analysis on the 2 main hyperparameters of the fine-tuning process.
\item To facilitate related communities, all related codes and resources are open-sourced in our Github repository\footnote{\href{https://github.com/SakiRinn/LLM4CVD}{https://github.com/SakiRinn/LLM4CVD}} and HuggingFace repository\footnote{\href{https://huggingface.co/datasets/xuefen/VulResource}{https://huggingface.co/datasets/xuefen/VulResource}} for more convenient reproduction.

\end{itemize}

The remainder of this paper is organized as below.
Section \ref{sec:RW} discusses the CVD task and three kinds of model architecture to tackle this task. 
Section \ref{sec:motivation} states our motivation to carry out this work. 
Section \ref{sec:prelin} introduces the problem definition and related preliminary knowledge. 
Section \ref{sec:bench} elaborates on our evaluated models and pre-processed experimental datasets. 
Section \ref{sec:exp} showcases the experimental results and related findings. 
Section \ref{sec:conclusion} summarizes this study, then discusses the limitation of this work and potential future directions.

\section{Related Works}
\label{sec:RW}
\textbf{Code Vulnerability Detection (CVD).}
Code vulnerability detection (CVD) serves as a significant role in the secure software systems. 
Previous CVD methods can be mainly divided into the \textit{dynamic approach} and the \textit{static approach} \cite{CVDsurvey}. 
For the \textit{dynamic approach}, representative methods like fuzzing testing technique \cite{fuzzy}  aim to identify code vulnerabilities by executing code programs, and observing the program output or internal states, which often leads to more human expertise and efforts. 
For the \textit{static approach}, representative methods aim to analyze code vulnerability without putting the code into the run time. 
Deep learning models mainly belong to the static approach, which have become mainstream research direction in recent years. These models are expected to analyze the code context and predict its vulnerability with minimum human efforts.
Herein we mainly discuss some featured deep learning models, and we roughly divide them into three groups including graph-based models, medium sequence models and pre-trained large language models.

\textbf{Graph-based models.}
Early attempts to perform CVD tasks basically exploit graph neural networks (GNN) \cite{gnnn} to identify vulnerabilities. 
Given a code instance, the general pipeline of a graph-based model constructs a code graph to represent the code, optimizes the embedding vector of the graph, and classifies the vector as vulnerable or non-vulnerable.
The graph can be formulated by Abstract Syntax Tree (AST), Control Flow Graph (CFG), Data Flow Graph (DFG), Program Dependence Graph (PDG), code property graph (CPG \cite{cpg}) or other formats, as introduced in \cite{survey}. 
ReVeal \cite{reveal} constructs the CPG and uses features obtained from this CPG.
VulChecker \cite{vulchecker} proposes a new enriched PDG format and idenitifies the vulnerability.
Devign \cite{devign} constructs a CPG and designs a novel convolutional module that can extract useful features from the learned node representation for graph-level classification.
ReGVD \cite{regvd} exploits two graph construction methods to encode its code graph with nodes representing code tokens and features initialized based on CodeBERT's code token embedding \cite{codebert}.

\begin{figure*}[htbp]
\begin{center}
\includegraphics[width=1.0\textwidth]{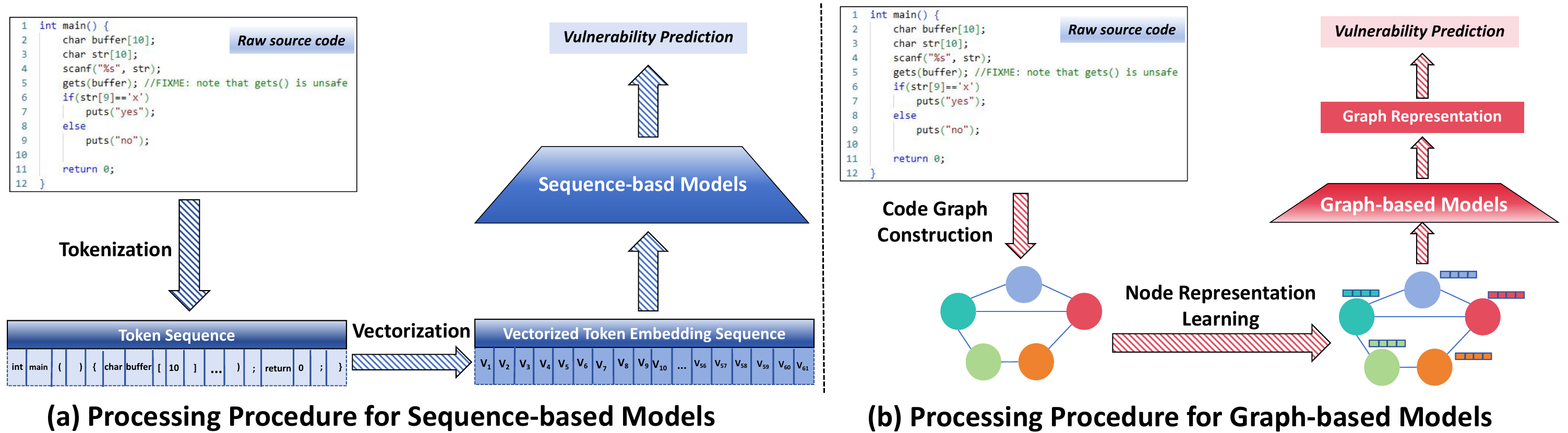}
\caption{Processing procedures for sequence-based models and graph-based models. We use simple naive tokenizer in this figure as an illustrative example.}
\label{fig:pipeline}
\end{center}
\end{figure*}

\textbf{Medium-size sequence models.}
Some early attempts like VulDeePecker \cite{vuldeepecker} and SeSyVR \cite{SeSyVR} aim to identify code vulnerability via light-weight sequence models like TextCNN, RNN or LSTM \cite{regvd,rnnn}. However, these early small-size sequence models are quickly surpassed by pre-trained Transformer \cite{transformer} based language models which are widely trained on a large corpus \cite{tdsc}.  
The parallel processing ability of these traditional sequence models is also limited, and it is also difficult to capture the association between long-distance tokens. 
Meanwhile, Transformer-based pre-trained language models \cite{transformer,bert} (or code pre-trained models as referred in \cite{vulllm}) have better scalability to the input context length than these early sequence models.
Achieving such input length scalability and long-distance token association is mainly credited to the attention mechanism \cite{transformer}.
These pre-trained language models often adopt a new learning paradigm of  'pre-training and fine-tuning', where the pre-training stage aims to learn general semantic information on large-scale corpus and the fine-tuning stage aims to relatively quickly adapt to the downstream CVD tasks \cite{stagedvulbert}. Representative models include CodeBERT \cite{codebert} and UniXcoder \cite{unixcoder}, which will be detailedly discussed in Section \ref{sec:pre-processing}

\textbf{Large language models (LLMs).} 
Compared with the above-discussed medium-size sequence models, LLMs can be regarded as large-scale sequence-based models since they have significantly larger parameter space and effectively undergo large-scale tremendous training corpus over trillions of tokens.
LLMs can be divided into close-source ones such as ChatGPT series \cite{chatgpt}, and open-source ones like Llama series \cite{Llama-2,Llama-3,codeLlama}. 
The model architecture of open-source LLMs is usually composed of multiple ($\textgreater$8) Transformer encoders or decoders.
LLMs have demonstrated impressive capabilities across diverse downstream tasks in recent studies, therefore, it is natural for code intelligence communities to leverage them for code-related tasks \cite{codeIntelligence}.
Much effort has been put into code generation \cite{codegen} and code repair \cite{coderepair}. There are already some famous programming assistants such as Copilot \cite{copilot}, CodeGeeX \cite{codegeex} and Cursor \cite{cursor}. These works are mostly generation-oriented tasks, while few studies aim to investigate the potential of these LLMs to predict whether a source code contains vulnerabilities \cite{xinzhou}. 
Some works \cite{xinzhou,zhangchao,nong} aim to fix the LLMs' weights and explore the effective prompt design to perform the CVD task. 
One recent work \cite{vulllm} firstly tries to fine-tune LLMs to predict the code vulnerabilities, but it misses evaluation on long code samples ($\textgreater$512 tokens) where code vulnerabilities often exist in \cite{stagedvulbert}.

\section{Motivation}
\label{sec:motivation}
With the joint efforts from software engineering, machine learning, natural language processing and other domains, there is a thriving achievement in code intelligence community \cite{codeIntelligence}. Code vulnerability detection (CVD) is one of the key challenges, while there are not many studies that focus on the potential of exploiting large language models (LLMs) for this challenge. To this end, we revisit the most recent related literature and propose this experimental study. Compared with existing works, our motivation is briefly two-fold:

\textbf{Unified Evaluation.} 
Gao et. al. propose VulBench \cite{zhangchao} to directly evaluate the LLMs' performance on the CVD task, which is an early attempt to explore LLMs' potential. 
Zhou et. al. \cite{xinzhou} propose to design different prompting templates to query the close-sourced ChatGPT.
Nong et. al. \cite{nong} propose to study specific prompting technique to query two open-source LLMs including Llama-2 \cite{Llama-2} and Falcon \cite{falcon}, and one close-source LLM ChatGPT \cite{chatgpt}.
Above studies aim to fix the model weights and explore the model performance with different prompting templates.
To our best knowledge, Du et. al. firstly propose VulLLM \cite{vulllm} to investigate the performance of fine-tuned LLMs for the CVD task, however, they miss to investigate long and complex code programs ($\textgreater$512 tokens). 
One recent study \cite{stagedvulbert} points out vulnerabilities often exist in these long programs, which is also in accordance with our statistics in Table \ref{tab:datasets}.
We find these works investigate models' performance on un-unified CVD datasets, which motivates us to carry out the unified evaluation on five relatively more-commonly utilized CVD datasets which cover both short code samples and long code samples.

\textbf{Unified and Easy-to-use Open-source Implementation.} In addition, during we carry out this study, we find there lacks a unified open-source codebase to train and evaluate both graph-based models, medium-size sequence models and LLMs, which brings obstacles for related communities to carry out re-implementation. 
Therefore, based on their open-source Github or HuggingFace repositories listed in Section \ref{sec:processing}, we implement nine related models as shown in Table \ref{tab:models} and carefully integrate them into one unified codebase to better facilitate related communities.
For LLMs, we investigate two advanced Llama series (Llama-3 and Llama-3.1 \cite{Llama-3}) which are rarely studied in previous CVD works.
Meanwhile, we provide the five most commonly used pre-processed datasets with a unified format. We organize all training or fine-tuning codes in an easy-to-use manner, which make it easier for re-implementations of graph-based models, medium-size sequence models and LLMs.

\section{Preliminaries}
\label{sec:prelin}
\subsection{Problem Definition}

In general, code vulnerability detection (CVD) is often formalized as a binary classification problem, i.e., predicting whether a given raw source code is vulnerable \cite{survey}. 
We define a vulnerable code dataset as  $( (c_{i},y_{i})|c_{i}\in\mathcal{C},y_{i}\in\mathcal{Y}), i\in\{1,2,\ldots,n\}$, where $\mathcal{C}$ denotes the set of $n$ code samples, ${\mathcal{Y}}=\{0,1\}^{n}$ denotes the label set where 1 and 0 represent the vulnerable code and benign code. The optimization objective for a model is to learn a mapping from $\mathcal{C}$ to $\mathcal{Y}$ denoted as $f\colon{\mathcal{C}}\mapsto{\mathcal{Y}}$ to estimate a code is vulnerable or not, and $f$ is expressed by a deep neural network. The optimization objective can be formed as 
\begin{equation}
    \min\sum_{i=1}^n\mathscr{L}\left(f\left(c_i,y_i|c_i\right)\right)+\lambda\omega(f),
\end{equation}
where $\mathscr{L}$ denotes the loss function for classification, $\omega(f)$ denotes the weight regularization \cite{l2decay} and $\lambda$ denotes the trade-off coefficient.  $f$ can be implemented by sequence models or graph-based models.

\subsection{Sequence Models}

In deep learning-based code vulnerability detection, code is typically represented as sequences or graph structures, serving as foundational inputs for neural network models. These representation methods encapsulate code semantics, enabling models to perform context-aware vulnerability analysis. 

Sequence models are generally paired with a tokenizer that transforms source code into token sequences, enabling the models to process them for vector representation generation. After tokenization, the sequence model applies embedding techniques, such as Bag of Words or Word2Vec, to convert tokens into vectors, which is called vectorization. The vectorized sequence is then inputted into the model’s main architecture for further forward computing. The sequence model ultimately outputs a vulnerability prediction, indicating whether the code is vulnerable or not vulnerable (i.e. benign), as illustrated in Figure \ref{fig:pipeline}(a).
Each code sample $c_i$ contains a relatively long word sequence.
Early sequence models utilize simple word-level tokenzier.
Bert-based models utilize the WordPiece as the tokenizer \cite{codebert}.
Large language models (LLMs) can be regarded as large-scale sequence models, and most LLMs utilize Byte-Pair Encoding (BPE) technique \cite{bpe} as their tokenizer.

\subsection{Instruction Tuning}

Instruction tuning aims to optimize the response of LLMs to specific instructions, thus ensuring the alignment with the requirements of a specific given task. 
Detailedly, we employ instruction tuning to fine-tune LLMs for code vulnerability detection task. 
By integrating this instruction with the input code, fine-tuned LLMs are capable of producing specific outputs. Subsequently, the LLM quantifies the discrepancy between the generated output and the anticipated target, leveraging this deviation to fine-tune the weights of LLM. In this work, we adapt the template provided by Alpaca \cite{alpaca}.

\begin{figure}[htbp]
\begin{center}
\includegraphics[width=0.5\textwidth]{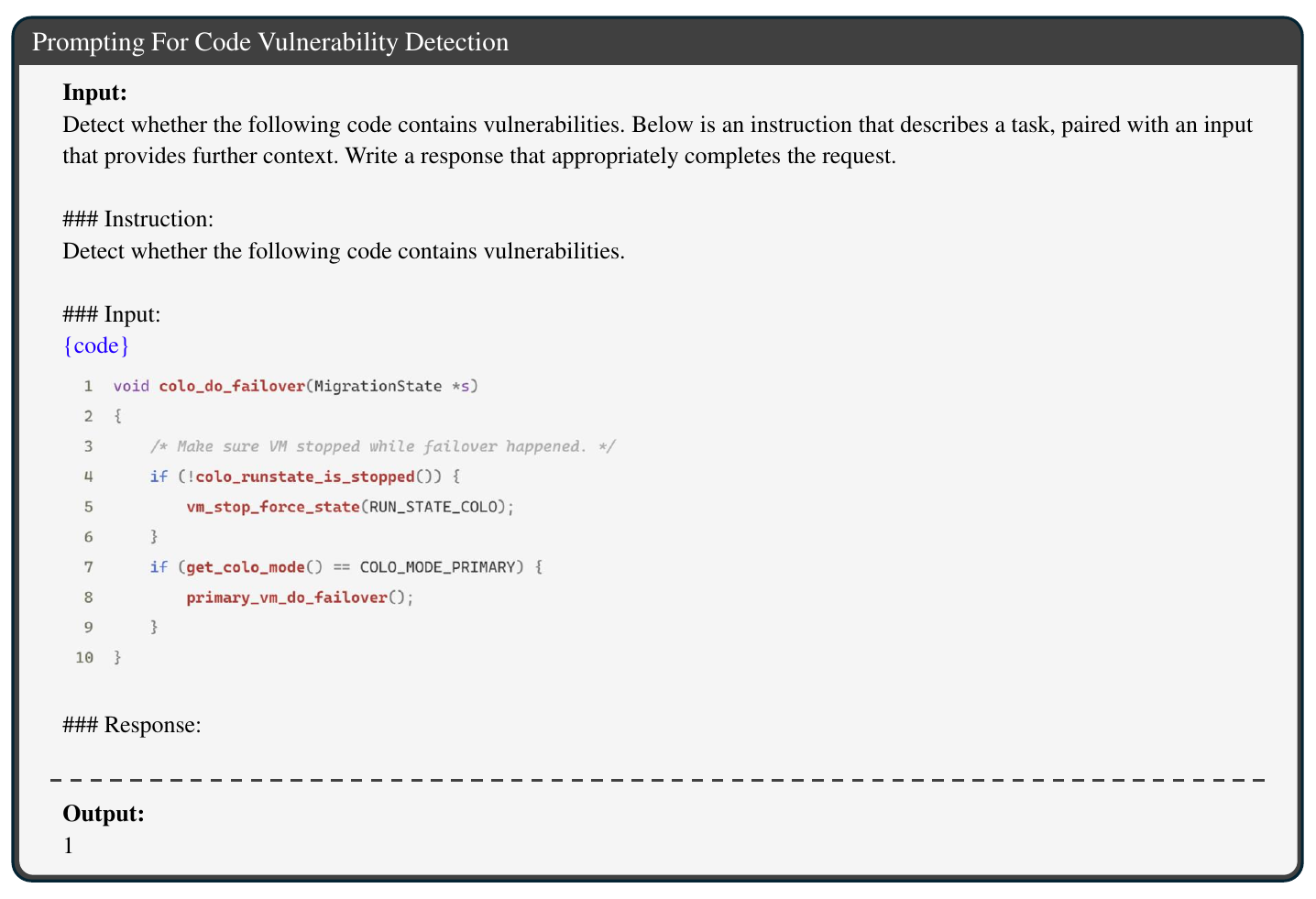}
\caption{Prompt template for large language models. \{code\} indicates the code content to be filled in. }
\label{fig:template}
\end{center}
\end{figure}

In detail, we use the most popular light-weight fine-tuning method Low-Rank Adaptation (LoRA \cite{lora}) to fine-tune four evaluated open-source LLMs (i.e. Llama-2, CodeLlama, Llama-3, and Llama-3.1).
The key idea of LoRA is to freeze the pre-trained model's weights and introduce trainable low-rank matrices as extra model bypass branches. These matrices are used to capture the task-specific adaptations. 
By doing so, it effectively reduces the number of trainable parameters and speeds up training while effectively adapting the model to new tasks in a specific domain.
Following \cite{vulllm}, the target modules to fine-tune are set to $q_{proj}$, $v_{proj}$, $k_{proj}$, and $o_{proj}$ in Self-attention layers \cite{transformer}.

\section{Benchmark Design}
\label{sec:bench}

\subsection{Evaluated Models}
\label{sec:processing}
Herein we elaborate on the detailed methodology of evaluated methods, which cover both classic deep learning (DL) based models and fine-tuning large language models (LLM).

\paragraph{Graph-based models} We choose three widely-used graph-based models. We refer to the training codes for Devign and ReGVD, which can be found in the following two Github repositories of Devign\footnote{\href{https://github.com/saikat107/Devign}{https://github.com/saikat107/Devign}} and ReGVD\footnote{\href{https://github.com/daiquocnguyen/GNN-ReGVD}{https://github.com/daiquocnguyen/GNN-ReGVD}}. The implementation of GraphCodeBERT is in accordance with other two medium-size sequence models which we will introduce later.

\begin{itemize}
    \item Devign \cite{devign} aims to encode a source code into a joint graph structure from multiple syntax and semantic representations and then leverage the composite graph-level representation to effectively learn to discover vulnerable code.
    \item ReGVD \cite{regvd}  encodes source code as a graph with nodes representing code tokens and features initialized based on pre-trained CodeBERT. The model combines sum and max pooling for graph-level embedding, which is then forwarded to a fully-connected and softmax layer to predict its vulnerabilities.
    \item GraphCodeBERT \cite{graphcodebert} is a new pre-trained programming language model, extending CodeBERT to consider the inherent structure of code data flow into the training objective. 
\end{itemize}

\paragraph{Medium-size Sequence models} We choose two widely-used medium-size code pre-trained models which are developed by Microsoft$^\circledR$, the training codes can be found in their Github repositories\footnote{\href{https://github.com/microsoft/CodeBERT}{https://github.com/microsoft/CodeBERT}}.

\begin{itemize}
    \item CodeBERT \cite{codebert} is a pre-trained model based on BERT for six programming languages (Python, Java, JavaScript, PHP, Ruby and Go), using masked language model \cite{bert} and replaced token detection \cite{replacedtoken} objectives during the pretraining process. Following the common practice \cite{codebert,dpccl, stagedvulbert}, the maximum input token length limit is fixed 512. Therefore, related experiments with CodeBERT or UniXcoder on long samples are not conducted, and neither reported in this study.
    \item UniXcoder \cite{unixcoder} leverages multi-view contents including the code abstract syntax tree (AST) and code comment to enhance code representation. It transforms the AST in a sequence structure that retains all structural information from the AST.
\end{itemize}

\paragraph{Large language models (LLMs)} We choose four widely-used LLMs which are all developed and free open-sourced by Meta AI $^\circledR$. The model checkpoints are provided in their HuggingFace repositories\footnote{\href{https://huggingface.co/meta-Llama}{https://huggingface.co/meta-Llama}}. The codes to fine-tune LLMs are referred from the Github repository of VulLLM\footnote{\href{https://github.com/CGCL-codes/VulLLM/tree/main/CodeLlama}{https://github.com/CGCL-codes/VulLLM/tree/main/CodeLlama}}.

\begin{itemize}
    \item Llama-2-7B \cite{Llama-2} is designed for a wide range of NLP tasks, including coding-related activities. It is currently one of the most widely used open-source large language models. The Llama-2 series is one of the earliest open-source LLMs. 
    Fine-tuned Llama-2 models outperform most concurrent open-source models on benchmarks such as MMLU, and achieve performance comparable to closed-source models like GPT-3 \cite{chatgpt}.
    \item CodeLlama-7B \cite{codeLlama} is a code-specialized model based on Llama-2 which specializes in and enhances code generation capabilities while addressing limitations in handling long contexts and zero-shot instruction following. It is announced in August, 2023. 
    As a code-specialized large language model, Code Llama surpasses the performance of open-source models like Llama-2 on various code benchmarks.
    \item Llama-3-8B \cite{Llama-3} is announced in April 2024 and claimed to be a major leap over Llama-2-7B. Compared to the Llama-2 series, Llama-3 focuses on optimizing data, scale, and complexity management, significantly improving performance in tasks such as multilingual processing, coding, reasoning, and tool utilization. 
    \item Llama-3.1-8B \cite{Llama-3} is announced in July, 2024. It is claimed to get performance improvement with the assistance of more controllable and simple post-training techniques. 
\end{itemize}

\begin{table*}[htbp]
\centering
\small
\caption{Details on the evaluated models.}
\label{tab:models}
\begin{tabular}{c|c|c|c|c}
\toprule\toprule
\textbf{Model Arch.} &\textbf{Venue}& \textbf{Parameter Scale}& \textbf{Type}  & \textbf{Main Model Component(s)} \\ \midrule
Devign \cite{devign}  &NeurIPS'19& 1M & Graph & GNN, Convolutional Layer \\ 
ReGVD \cite{regvd} &IEEE ICSE'22& 125M& Graph & GNN, CodeBERT \\ 
GraphCodeBERT \cite{graphcodebert} &ICLR'21& 125M& Graph &Transformer Encoder \\ \midrule
CodeBERT \cite{codebert} &EMNLP'20& 125M& Sequence&  Transformer Encoder \\ 
UniXcoder \cite{unixcoder} &ACL'22& 126M& Sequence& Transformer  \\ \midrule
Llama-2-7B \cite{Llama-2} &Arxiv'23& 7B& Sequence & Transformer Decoder \\ 
CodeLlama-7B \cite{codeLlama} &Arxiv'23& 7B& Sequence & Transformer Decoder \\ 
Llama-3-8B \cite{Llama-3} &Arxiv'24& 8B& Sequence & Transformer Decoder \\ 
Llama-3.1-8B \cite{Llama-3} &Arxiv'24& 8B& Sequence & Transformer Decoder  \\ 

\bottomrule
\end{tabular}
\end{table*}

\subsection{CVD Datasets}
\label{sec:datasets_detail}

In the communities of code vulnerability detection, there are several existing previous works that curate code datasets containing both benign code samples and vulnerable ones. In this study, we refer to related literature \cite{zhangchao, vulllm, devign, reveal} and select commonly-used C/C++ function-level datasets for experiments.

\begin{itemize}
    \item ReVeal \cite{reveal} is labeled using the patches to known security issues at Chromium security issues and Debian security tracker. ReVeal considers the changed functions before a security patch (commit) as vulnerable, after the patch as non-vulnerable, and all unchanged functions as non-vulnerable. 
    \item Devign \cite{devign} dataset is firstly created by Zhou et al, \cite{devign}, including 27,318 manually labeled vulnerable or non-vulnerable functions extracted from security-related Github commits in two large and popular C programming language open-source projects (i.e. QEMU and FFmpeg) and diversified in functionality \cite{survey}. Devign has high-quality labels since it is annotated by three security experts, but manual labeling is very expensive, which costs around 600 man-hours.
    \item Draper \cite{draper} dataset generated labels by selecting the alert categories from three static analyzers: Clang, Cppcheck, and Flawfinder. It includes millions of C/C++ function-level examples collected from the SATE IV Juliet test suite, Debian Linux, and GitHub repositories with some synthesized samples. All samples are normalized using a custom C/C++ lexer, removing redundant information such as code comments, and are deduplicated to ensure data quality. The quality of the label is unknown and less investigated, but the label accuracy of static analyzers tends to be low as reported in \cite{diversevul}.
    \item BigVul  \cite{bigvul} collects vulnerability fixing commits from Common Vulnerabilities and Exposures (CVE) entries from 348 projects \cite{stagedvulbert,bigvul}, covering 3,754 code vulnerabilities among 91 vulnerability types. BigVul performs a preliminary search by using automated tools to filter C/C++ projects on GitHub, detecting commits that might be linked to vulnerabilities. These commits are then cross-checked using bug reports and matched to CVE entries.
    \item DiverseVul \cite{diversevul} stands out for its diversity. It collects 7,514 commits from 797 projects and covers up to 150 CWE vulnerability types. Its collection methodology is similar to the ReVeal dataset, marking the before-commit version of a function as vulnerable and the rest as benign, with deduplication performed using the MD5 hash of functions. Finally, all vulnerable functions are manually mapped to corresponding CVE and CWE entries.
\end{itemize}

We summarize related statistics of these datasets in Table \ref{tab:datasets}. Except Devign \cite{devign}, other datasets exhibit obvious class imbalance. We subsample part of the samples in Draper, BigVul and diverseVul. More details regarding our pre-processing procedures can be found in Section \ref{sec:pre-processing}.

Meanwhile, providing a high-quality annotated code dataset is expensive, so some datasets like Draper and D2A \cite{d2a} contain non-negligible noisy labels \cite{nll,fnbench} as discussed in previous studies \cite{diversevul}. Among the datasets we selected, only Devign explicitly states that data annotation is performed by security experts, ensuring high data quality. The other datasets rely solely on auto-labelers, security patches, and commits for annotation, which raises concerns about low data quality and incompleteness. To cope with the underlying label noise in CVD datasets, we leave it as our future works.

\begin{table*}[htbp]
\centering
\caption{Details on the evaluated CVD datasets. Vul. Ratio indicates the proportion of the vulnerable samples across the samples ($\sim$50\% indicates a relatively balanced dataset). The number of samples before subsampling is indicated in parenthesis; see Section \ref{sec:pre-processing} for details. }
\label{tab:datasets}
\resizebox{\linewidth}{!}{
\begin{tabular}{c|c|c|c|c|c|c|c}
\toprule \toprule
\textbf{Dataset} & \textbf{\begin{tabular}[c]{@{}c@{}}Short\\ Samples\end{tabular}}  &\textbf{\begin{tabular}[c]{@{}c@{}}Vul. Ratio of \\ Short Samples\end{tabular}} & \textbf{\begin{tabular}[c]{@{}c@{}}Long\\ Samples\end{tabular}}  &\textbf{\begin{tabular}[c]{@{}c@{}}Vul. Ratio of \\ Long Samples\end{tabular}} & \textbf{Total} &\textbf{\begin{tabular}[c]{@{}c@{}}Annotation\\ Method\end{tabular}}              & \textbf{\begin{tabular}[c]{@{}c@{}}Sample\\ Type\end{tabular}} \\ \midrule
ReVeal \cite{reveal}           &                                                                  
18,387 &6.90\%
&                                                                 2,456 &18.57\%
&                20,843&Security Issues                                                              & Real-world                                                             \\ \midrule
Devign \cite{devign}          &                                                                  19,221 &44.08\%
&                                                                 4,529 &48.82\%
&                23,750&Labeled by Experts    & Real-world      \\ \midrule
Draper \cite{draper}          &                                                                  \begin{tabular}[c]{@{}c@{}}25,000\\ (1,147,893)\end{tabular} &5.80\%
&                                                                 \begin{tabular}[c]{@{}c@{}}2,262\\ (122,247)\end{tabular} &12.55\%
&                \begin{tabular}[c]{@{}c@{}}27,662\\ (1,270,140)\end{tabular}&\begin{tabular}[c]{@{}c@{}}Stable Analyzer\\ \& Category Filter\end{tabular} & \begin{tabular}[c]{@{}c@{}}Real-world\\ \& Synthetic\end{tabular}                                                           \\ \midrule

BigVul \cite{bigvul}          &                                                                  \begin{tabular}[c]{@{}c@{}}25,000\\ (168,605)\end{tabular} &4.46\%
&                                                                 \begin{tabular}[c]{@{}c@{}}1,882\\ (12,694)\end{tabular} &12.33\%
&                \begin{tabular}[c]{@{}c@{}}26,882\\ (181,299)\end{tabular}&Security Issues                                                              & Real-world                                                             \\ \midrule
 DiverseVul \cite{diversevul}& \begin{tabular}[c]{@{}c@{}}25,000\\ (273,785)\end{tabular} &3.94\%& \begin{tabular}[c]{@{}c@{}}3,039\\ (33,274)\end{tabular} &10.79\%& \begin{tabular}[c]{@{}c@{}}28,039\\ (307,059)\end{tabular} &Security Issues                                                              & Real-world                                                             \\  \bottomrule
\end{tabular}
}
\end{table*}

\subsection{Dataset Pre-processing}
\label{sec:pre-processing}

In this study, we focus on identifying key factors during training that influence the fine-tuned LLM's detection performance. The dataset serves as the cornerstone of fine-tuning LLMs, as different datasets can lead to vastly divergent outcomes, making it undoubtedly the most critical component of fine-tuning. However, many existing studies' benchmarks solely involve only 1–2 datasets \cite{xinzhou,stagedvulbert,regvd}, making it obscure to comprehensively evaluate a model's detection performance across various scenarios.

As fine-grained statistics listed in Section \ref{sec:datasets_detail}, our study involves 5 influential and widely-used datasets in this field. 
This enables us to observe how well each model performs when confronted with various types of vulnerabilities. 
Due to the significant number of involved datasets and the obvious differences in attributes such as sample size and positive sample ratios, we applied the following data pre-processing steps in the main experiments to ensure fairness in evaluation:

\begin{itemize}
\item \textbf{Filtering.} As mentioned in Section \ref{sec:datasets_detail}, the quality of code datasets varies significantly. We find anomalies in some samples during data preprocessing. In the DiverseVul dataset, We are unable to trace some samples based on their `project' and `commit\_id' attributes. In the Draper dataset, annotation inaccuracies are prevalent, particularly in code samples associated with multiple CWE types, where obvious labeling errors are found. To address this, we perform an initial filtering of these two datasets to exclude anomalous samples. Note that the data quality and label noise issues are also pointed out in previous works \cite{diversevul}, which leaves space for future works.
\item \textbf{Formatting.}
Diverse representations and storage formats of samples pose challenges for conducting unified experiments. We format every dataset in order to avoid this. We assign a unique index to each sample and used the `code' and `label' attributes to represent every sample's code and label respectively. Additionally, we retain additional attributes specific to each dataset, such as CWE type and commit ID, which can assist with future works like vulnerability line extraction and vulnerability classification. Notably, only the `code' and `label' attributes are used in all of our experiments in Section \ref{sec:exp}. 
To facilitate the fine-tuning of LLMs to adapt the CVD classification task, the labels are annotated to $\mathbf{1}$ or $\mathbf{0}$ to denote the vulnerable and benign class, following the previous successful practice \cite{drivelm,vulllm}. Data are formatted using the general instruction fine-tuning template provided by Alpaca \cite{alpaca} format, as illustrated in Figure \ref{fig:pipeline}. In this template, [Input] and [Output] are derived from the aforementioned data preparation process, while [Task Prompt] guides the LLM to generate task-specific outputs based on different tasks.
\item \textbf{Division by Sequence Length.}
Some code pre-trained models, such as CodeBERT \cite{codebert} and UniXcoder \cite{unixcoder}, include learnable positional encodings, which constrain the input sequence length to 512 tokens \cite{stagedvulbert}. Extending positional encodings beyond this limit requires reinitializing the extended encodings, making it impossible to leverage pre-trained parameters fully. This could result in unpredictable performance degradation. To address this, we divide the datasets into short and long samples, using a sequence length of 512 as the boundary.
We serialize all dataset samples using Llama-3 tokenizer \cite{Llama-3}, which is one of the most advanced tokenizers, and calculate the sequence lengths to divide each dataset into long-sample and short-sample subsets, following the practice of VulLLM \cite{vulllm}.
Thus, each dataset has two subsets, where one subset contains \textbf{short samples} with lengths less than 512 and another subset contains \textbf{long samples} with lengths between 512 and 1024. Samples containing more than 1024 tokens are excluded due to their large variation in length (some even exceeding 10K tokens). The resource cost of training and inference these extra long samples would be unaffordable.
\item \textbf{Subsampling.}
There are significant differences in the number of samples across certain datasets. For example, the number of samples in Draper is more than 50 times that of ReVeal. Excessively large datasets extremely increase training costs and create imbalances that introduce implicit biases to the model. To address this, we applied subsampling to the datasets. 
We subsample part of samples in Draper \cite{draper}, BigVul \cite{bigvul} and DiverseVul \cite{diversevul}. Since each dataset is divided into long-sample and short-sample subsets, and short-sample subsset typically contain far more samples than the long-sample ones, we applied subsampling to the short-sample datasets, limiting the maximum number of samples to 25,000. Then, we apply the same proportional rate of subsampling to the long-sample datasets as we do to the corresponding short-sample datasets. All pre-process procedure codes are released for reference.
\end{itemize}

\section{Experiments}
\label{sec:exp}
\subsection{Experimental Settings}

For experiment, we investigate the LLMs' performance compared with graph-based models and medium-size sequence models. Herein we introduce related experimental settings for implementation.

\textbf{Environments} All the experiments are conducted on an Ubuntu 20.04 server with AMD$^\circledR$ Ryzen 24-Core Processor CPU, and 1 NVIDIA$^\circledR$ L20 GPU (48G). The computational backend is PyTorch 2.1.0 and CUDA 12.1.

\textbf{Datasets} We conduct experiments on 5 widely-used code vulnerability datasets as elaborated in Section \ref{sec:datasets_detail}. We divided each dataset into two subsets (long samples and short samples) based on the sample length. Subsequently, each subset is split into train, validation, and test sets in the ratio of 8:1:1.
Graph models cannot directly process sequential data. 
Therefore, we used Joern \cite{joern}, a CPG \cite{cpg} based C/C++ code analysis tool, to convert each sequence sample into a code graph, which is aligned with the processing methods of Devign \cite{devign} and ReVeal \cite{reveal}. 
To obtain feature vectors for each node in the graph, we trained a Word2Vec \cite{word2vec} model with a vector size of 200 to vectorize each token. The converted graph dataset is stored in JSON format.

\textbf{Models and Hyper-parameters} We evaluate 3 graph-based models, 2 medium-size sequence models and 4 large language models (LLMs). We provide the Github repositories for all implementations of fine-tuning LLMs and baseline models in Section \ref{sec:processing}.
For the classical graph-based model Devign\cite{devign}, the input feature size and graph embedding size are set to 200. 
Adam is used as the optimizer with a learning rate of 1e-4 and a weight decay of 1e-3.
Both of medium-size sequence models and all graph-based models except Devign are methods based on Transformer encoder, and their hyperparameter settings are consistent. 
The block size of them is set to 512 for short sample datasets and 1024 for long sample datasets. AdamW is used as the optimizer with a learning rate of 2e-5.
For fine-tuning LLMs, we configure the model parameters using the default settings provided by Meta AI$^\circledR$. We employ LoRA\cite{lora} for fine-tuning, setting the rank to 16, the scaling factor $\alpha$ to 32, and dropout rate to 0.05. AdamW is used as the optimizer with a learning rate of 1e-4, and the model is trained for 5 epoch. We fine-tune the $q_{proj}$, $v_{proj}$, $k_{proj}$, and $o_{proj}$ weight matrices in the self-attention layers following \cite{vulllm}.

\textbf{Metrics} 
To evaluate the performance of our proposed method, we use the following five metrics computed by the confusion matrix\footnote{The confusion matrix contains 4 components including true positive (TP), true negative (TN), false negative (FN), and false positive (FP) \cite{f1score}.}, which have been widely accepted by previous work \cite{stagedvulbert,SeSyVR}:
\begin{itemize}
    \item Acc. : Accuracy (Acc.) is a widely used metric for a classification task, which can be calculated by $Acc.=(TP+TN)/(TP+ FP+FN+TN)$.
    \item Pre. : Precision (Pre.) rate is the fraction of predicted vulnerabilities that are correctly predicted: $Pre.= TP / (TP+FP)$.
    \item Rec. : Recall (Rec.) rate is the fraction of true positive vulnerabilities in the actual vulnerabilities: $Rec.= TP / (TP + FN)$.
    \item F1-Score: F1-Score denotes the harmonic mean of precision and recall and is calculated as: $F1 = 2\times(Pre. \times Rec.)/(Pre. + Rec.)$.  
    \item FPR: Referring to previous works \cite{diversevul}, we additionally utilize the false positive rete (i.e. FPR) as one metric since it reflects the probability that a negative sample is wrongly classified as a positive sample in a classification or detection system. It can be calculated by $FPR = FP/(FP + TN)$. 
\end{itemize}

$\quad$In highly imbalanced CVD datasets as introduced in Section \ref{sec:datasets_detail}, the commonly used accuracy (Acc.) metric yields misleadingly high performances that result from systematically predicting the majority class \cite{f1score}. Therefore, F1-Score can be more precise for the CVD task which is our main technical metric. In addition, FPR can assist to understand why a given classification system underperforms.

\begin{table*}[htbp]
\caption{Main experimental metrics (\%). We use the F1-Score as the main analyzed metric. The \textbf{bold} denotes the best result on this dataset while the \underline{underlined} denotes the second place result on this dataset. \textbf{-} denotes we do not conduct related experiments because these models do not support samples with more than 512 tokens as discussed in Section \ref{sec:processing}.}
\label{tab:main}
\small
\resizebox{\linewidth}{!}{
\begin{tabular}{c|c|ccccc|ccccc}
\toprule\toprule
\multirow{2}{*}{\textbf{Dataset}}   & \multirow{2}{*}{\textbf{Model Arch.}}     & \multicolumn{5}{c|}{\textbf{Short Samples}}   & \multicolumn{5}{c}{\textbf{Long Samples}} \\ 
 &  & \textbf{Acc. $\uparrow$}  & \textbf{Pre. $\uparrow$}  & \textbf{Rec. $\uparrow$}  & \textbf{F1-Score $\uparrow$}  & \textbf{FPR $\downarrow$}     & \textbf{Acc. $\uparrow$}  & \textbf{Pre. $\uparrow$}  & \textbf{Rec. $\uparrow$}  & \textbf{F1-Score $\uparrow$}  & \textbf{FPR $\downarrow$} \\ 
\midrule

\multirow{9}{*}{ReVeal \cite{reveal}} & Devign \cite{devign}                &92.06  &27.27  &12.40  &17.05              &2.33   &73.17  &9.52   &4.08   &5.71   &9.64   \\ 
& ReGVD \cite{regvd}                  &93.42  &0.00   &0.00   &0.00               &0.00   &80.08  &0.00   &0.00   &0.00   &0.00   \\ 
& GraphCodeBERT \cite{graphcodebert}  &93.69  &100.00 &4.13   &7.94               &0.00   &-&-&-&-&-\\ 
& CodeBERT \cite{codebert}            &92.82  &43.53  &30.58  &\underline{35.92}  &2.79   &- &- &- &- &-  \\ 
                                      & UniXcoder \cite{unixcoder}          &94.02  &59.32  &28.93  &\textbf{38.89}     &1.40   &- &- &- &- &-  \\ 
                                      & Llama-2-7B \cite{Llama-2}           & 93.15& 38.10& 6.61& 11.27& 0.76&                        77.24&                         41.03&                      32.65&                        \underline{36.36}&                   11.68\\ 
                                      & CodeLlama-7B \cite{codeLlama}      & 93.09& 36.36& 6.61& 11.19& 0.81&                        69.51&                         32.43&                      48.98&                        \textbf{39.02}&                   25.38\\ 
                                      & Llama-3-8B \cite{Llama-3}          & 92.33& 34.38& 18.18& 23.78& 2.44&                        75.61&                         32.26&                      20.41&                        25.00&                   10.66\\ 
                                      & Llama-3.1-8B \cite{Llama-3}      & 92.71& 36.17& 14.05& 20.24& 1.75&                        80.49&                         55.56&                      10.20&                        17.24&                   2.03\\ \midrule
\multirow{9}{*}{Devign \cite{devign}} & Devign \cite{devign}&                        52.52&                         48.64&                      79.98&                        60.49&                   70.35&                        52.32&                         51.88&                      66.96&                        \underline{58.46}&                   62.39\\ 
& ReGVD \cite{regvd}                 &                        56.94&                         52.80&                      49.66 &                        51.18&                   36.99&                        49.01&                         47.44&                      16.30&                        24.26&                   18.14\\ 
& GraphCodeBERT \cite{graphcodebert} &                        64.64&                         64.83&                      48.51&                        55.50&                   21.93&                        -&                         -&                      -&                        -&                   -\\ 
& CodeBERT \cite{codebert}           &                        64.85&                         66.28&                      46.11&                        54.39&                   19.54&                        -&                         -&                      -&                        -&                   -\\ 
                                      & UniXcoder \cite{unixcoder}         &                        65.63&                         60.35&                      71.05&                        \textbf{65.27}&                   38.89&                        -&                         -&                      -&                        -&                   -\\ 
                                      & Llama-2-7B \cite{Llama-2}           & 63.29& 67.50& 37.07& 47.86& 14.87&                        52.54&                         52.73&                      51.10&                        51.90&                   46.02\\ 
                                      & CodeLlama-7B \cite{codeLlama}      & 68.07& 73.99& 45.88& 56.64& 13.44&                        58.28&                         66.10&                      34.36&                        45.22&                   17.70\\ 
                                      & Llama-3-8B \cite{Llama-3}          & 67.65& 74.80& 43.48& 54.99& 12.20&                        53.42&                         52.82&                      66.08&                        \textbf{58.71}&                   59.29\\ 
                                      & Llama-3.1-8B \cite{Llama-3}      & 64.95& 61.42& 61.56& \underline{61.49}& 32.22&                        55.63&                         55.65&                      56.39&                        56.02&                   45.13\\ \midrule
\multirow{9}{*}{Draper \cite{draper}} & Devign \cite{devign}&                        92.72&                         23.81&                      14.49&                        18.02&                   2.71&                        87.27&                         27.78&                      19.23&                        \underline{22.73}&                   5.39\\ 
& ReGVD \cite{regvd}                 &                        94.48&                         0.00&                      0.00&                        0.00&                   0.00&                        90.26&                         0.00&                      0.00&                        0.00&                   0.00\\ 
& GraphCodeBERT \cite{graphcodebert} &                        93.48&                         38.94 &                      31.88 &                        35.06&                   2.92&                        -&                         -&                      -&                        -&                   -\\ 
& CodeBERT \cite{codebert}           &                        93.44  &                         39.34 &                      34.78&                        \underline{36.92}&                   3.13&                        -&                         -&                      -&                        -&                   -\\ 
                                      & UniXcoder \cite{unixcoder}         &                        92.72   &                         35.53&                      39.13&                        \textbf{37.24}&                   4.15&                        -&                         -&                      -&                        -&                   -\\ 
                                      & Llama-2-7B \cite{Llama-2}           & 94.36& 45.16& 10.14& 16.57& 0.72&                        90.64&                         60.00&                      11.54 &                        19.35&                   0.83\\ 
                                      & CodeLlama-7B \cite{codeLlama}      & 93.92& 40.54& 21.74& 28.30& 1.86&                        91.01  &                         100.00 &                      7.69&                        14.29&                   0.00\\ 
                                      & Llama-3-8B \cite{Llama-3}          & 92.44& 33.33& 36.96& 35.05& 4.32&                        91.39&                         63.64&                      26.92&                        \textbf{37.84}&                   1.66\\ 
                                      & Llama-3.1-8B \cite{Llama-3}      & 93.72& 34.92& 15.94& 21.89& 1.74&                        88.39  &                         30.77&                      15.38&                        20.51&                   3.73\\ \midrule
\multirow{9}{*}{BigVul \cite{bigvul}} & Devign \cite{devign}&                        95.80&                         53.85&                      6.60&                        11.76&                   0.25&                        76.19&                         4.76&                      3.85&                        4.26&                   12.27\\ 
& ReGVD \cite{regvd}                 &                        95.76&                         0.00&                      0.00&                        0.00&                   0.00&                        86.24&                         0.00&                      0.00&                        0.00&                   0.00\\ 
& GraphCodeBERT \cite{graphcodebert} &                        95.80&                         52.63&                      9.43&                        16.00&                   0.38&                        -&                         -&                      -&                        -&                   -\\ 
& CodeBERT \cite{codebert}           &                        95.56&                         38.10&                      7.55&                        12.60&                   0.54&                        -&                         -&                      -&                        -&                   -\\  
                                      & UniXcoder \cite{unixcoder}         &                        95.80&                         53.85&                      6.60&                        11.76&                   0.25&                        -&                         -&                      -&                        -&                   -\\ 
                                      & Llama-2-7B \cite{Llama-2}           & 98.96& 92.55& 82.08& \textbf{87.00}& 0.29&                        97.88&                         92.31&                      92.31&                        92.31&                   1.23\\ 
                                      & CodeLlama-7B \cite{codeLlama}      & 98.56                  & 84.31                   & 81.13                & 82.69                  & 0.67&                        97.88  &                         92.31&                      92.31 &                        92.31&                   1.23\\ 
                                      & Llama-3-8B \cite{Llama-3}          & 98.64& 86.00& 81.13& \underline{83.50}& 0.58&                        98.94&                         92.86&                      100.00&                        \textbf{96.30}&                   1.23\\ 
                                      & Llama-3.1-8B \cite{Llama-3}      & 98.60& 92.77& 72.64                & 81.48& 0.25&                        98.41 &                         92.59 &                      96.15 &                        \underline{94.34} &                   1.23\\ \midrule
\multirow{9}{*}{DiverseVul \cite{diversevul}} & Devign \cite{devign}&                        94.16&                         11.86&                      6.93&                        8.75&                   2.17&                        88.49&                         28.57&                      13.79&                        \underline{18.60}&                   3.64\\ 
& ReGVD \cite{regvd}                 &                        95.96&                         0.00&                      0.00&                        0.00&                   0.00&                        90.46&                         0.00&                      0.00&                        0.00&                   0.00\\ 
& GraphCodeBERT \cite{graphcodebert} &                        96.00  &                         100.00 &                      0.99&                        1.96&                   0.00&                        -&                         -&                      -&                        -&                   -\\ 
& CodeBERT \cite{codebert}           &                        95.84 &                         40.00 &                      5.94 &                        \underline{10.34}&                   0.38&                        -&                         -&                      -&                        -&                   -\\ 
                                      & UniXcoder \cite{unixcoder}         &                        95.64 &                         35.71 &                      9.90 &                        \textbf{15.50}&                   0.75&                        -&                         -&                      -&                        -&                   -\\ 
                                      & Llama-2-7B \cite{Llama-2}           & 95.96& 0.00& 0.00& 0.00& 0.00&                        89.47&                         20.00&                      3.45&                        5.88&                   1.45\\ 
                                      & CodeLlama-7B \cite{codeLlama}      & 95.84& 0.00& 0.00& 0.00& 0.13&                        90.13&                         0.00&                      0.00&                        0.00&                   0.36\\ 
                                      & Llama-3-8B \cite{Llama-3}          & 95.40& 20.83& 4.95& 8.00& 0.79&                        64.14&                         12.26&                      44.83 &                        \textbf{19.26}&                   33.82\\ 
                                      & Llama-3.1-8B \cite{Llama-3}      & 94.96& 16.22& 5.94& 8.70& 1.29&                        82.24&                         14.29&                      17.24&                        15.62&                   10.91\\ \bottomrule
\end{tabular}
}
\end{table*}

\subsection{Analysis on Main Experiments}
For main experiments, we train and evaluate 3 graph-based models, 2 medium-size sequence models and 4 LLMs on 5 widely-used code vulnerability datasets, as introduced in Section \ref{sec:bench}. Notably, only Devign is a relatively class balanced dataset, and other four datasets exhibit obvious class imbalance as shown in Table \ref{tab:datasets}. 

GraphCodeBERT, CodeBERT and UniXcoder cannot be evaluated on long sample datasets because they are all based on the RoBERTa \cite{roberta} architecture. The learnable position encoding layer of RoBERTa limits the input sequence length to 512 \cite{stagedvulbert}, meaning that long samples will be truncated to 512. Although ReGVD is also based on the RoBERTa architecture, it only uses the pretrained embedding layer and does not involve position encoding or any subsequent layers. Therefore, we can evaluate ReGVD on long sample datasets. Table \ref{tab:main} provides a detailed summary of the main experimental results.

\textit{Finding 1:} \textbf{The performance of all LLMs and other models tend to be influenced by the class imbalance of the dataset.} All models perform significantly better on the Devign dataset than on the other datasets, and Devign is also the most balanced dataset with nearly equal numbers of positive and negative samples. In contrast, all models tend to perform poorly on the DiverseVul dataset, which has the fewest positive samples. The difference between these two datasets is most clearly reflected in recall.
It is worth noting that in some experiments, all metrics except accuracy are 0. This phenomenon is most commonly observed on ReGVD, which only present normal metrics on the most balanced Devign dataset. Moreover, both Llama-2 and CodeLlama show this anomaly on the most imbalanced DiverseVul dataset. It emphasizes the important role of data balance.

\textit{Finding 2:} \textbf{The medium-size sequence models excel on short sample datasets, generally outperforming LLMs.} On the ReVeal, Draper, and DiverseVul, both of the medium-size sequence models achieve the highest and second-highest F1-scores respectively. There is an evident performance difference between the LLM and medium-sized sequence models. Although they have generally similar precisions, LLMs' recall rates are significantly lower than the medium-sized sequence models, resulting in lower F1- scores. This gap narrows as the dataset becomes more balanced, with the Devign dataset showing the smallest gap. We conclude that medium-sized sequence models are less affected by a low proportion of positive samples than LLMs. The code pre-trained Transformer encoder enables them to capture vulnerability features more accurately even with limited vulnerability data.

\textit{Finding 3:} \textbf{LLMs have potential to perform exceptionally well on long sample datasets.} 
Due to the limitation in parameter size, most CVD models have trouble to handle long samples effectively. Except for LLMs, only 2 of selected models can be evaluated on long sample datasets, and both performed far worse than the LLMs. LLM’s large parameter size and long context window ensure its  outstanding capacity to handle long samples. Additionally, for all the datasets we used, more of vulnerability samples are long samples, which could explain why the LLM generally performs better on long samples than on short samples in the same datasets. A larger number of vulnerability samples assists the LLM's learning of vulnerability features.

\textit{Finding 4:} \textbf{LLMs exhibit low FPRs, making it more reliable than other models.} 
FPR directly determines the reliability of a vulnerability detection tool\cite{fpr1}, and Excessive false positives (FPs) can hold developers from using the model in practice \cite{reveal, fpr2}. Except for the long sample part of ReVeal and DiverseVul datasets, LLMs have quite lower FPRs than other models without the compromise of overall performance. This advantage is particularly evident on short sample datasets.

From the above analysis, it is clear that the proportion of positive samples in the dataset plays a decisive role in the CVD performance of trained models, while the impact of sample length should not be overlooked. To specifically investigate the effects of positive sample ratio and length on fine-tuning  LLMs, we have designed experiments in Section \ref{sec:imbalance} and \ref{sec:length} respectively.

\begin{figure*}[htbp]
\begin{center}
\includegraphics[width=1.0\textwidth]{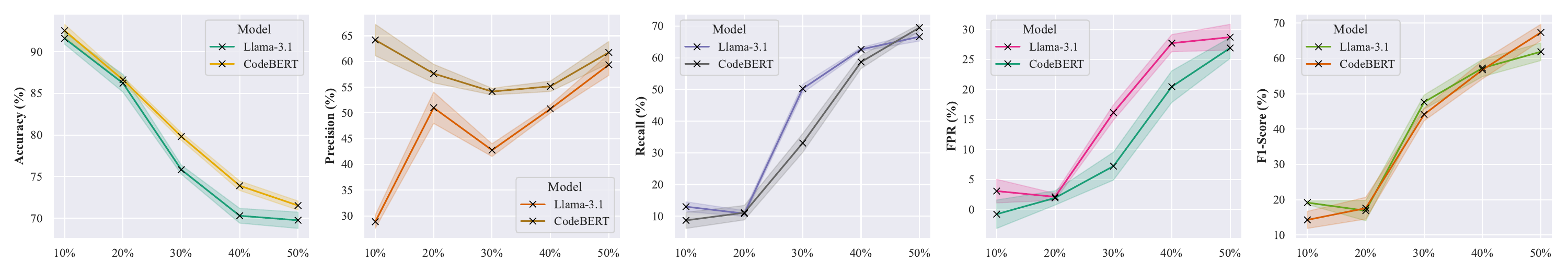}
\caption{Metrics on Varing Positive Sample Ratio on the DiverseVul \cite{diversevul} Dataset.}
\label{fig:pos ratio_diversevul}
\end{center}
\end{figure*}

\begin{figure*}[htbp]
\begin{center}
\includegraphics[width=1.0\textwidth]{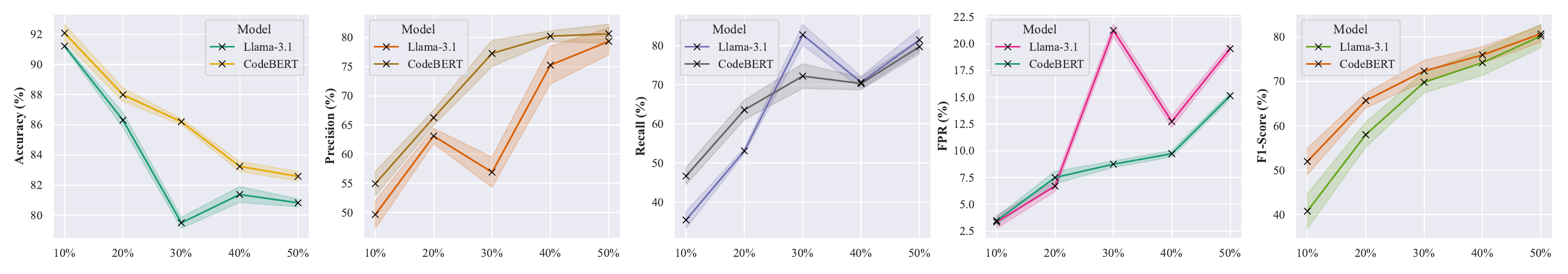}
\caption{Metrics on Varing Positive Sample Ratio on the Draper \cite{draper} Dataset.}
\label{fig:pos ratio_draper}
\end{center}
\end{figure*}

\begin{figure*}[htbp]
\begin{center}
\includegraphics[width=1.0\textwidth]{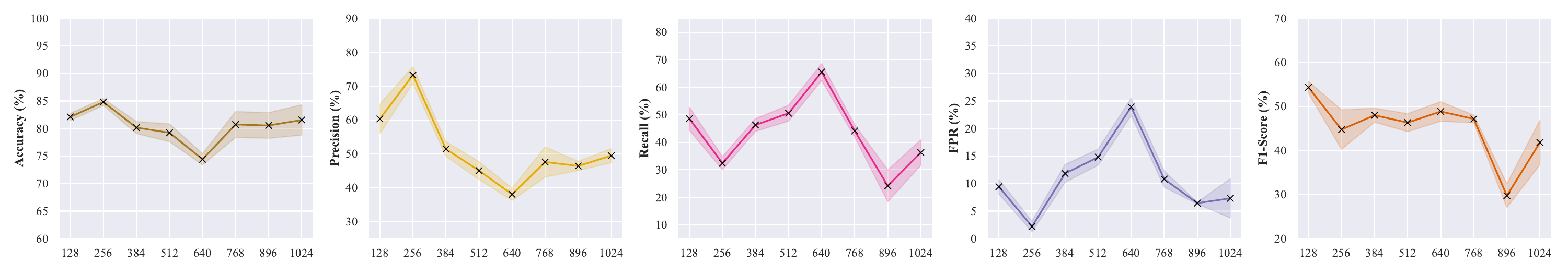}
\caption{Metrics on Varying Code Sequence Length.}
\label{fig:length}
\end{center}
\end{figure*}

\subsection{Analysis on Datasets with Varying Postive Sample Ratios}
\label{sec:imbalance}
As two recent study \cite{longtailed,xinzhou} and our statistics in Table \ref{tab:datasets} point out, long-tailed distribution within CVD datasets could pose a challenge for LLMs-based vulnerability detection solutions, and we can also observe this in our main experiments in Table \ref{tab:main}. Thus, we carry out the re-sampling experiments which creates more balanced datasets. 
We subsample the Draper and DiverseVul datasets \cite{draper,diversevul} to make the positive samples (i.e. vulnerable) to occupy more percentage across the training dataset, and the ratio is incrementally set to 10\%, 20\%, 30\%, 40\% and 50\%. Similar to the main experiment, the size of each sampled dataset is controlled at 25,000. We select CodeBERT as the studied medium-size sequence model and the LLama-3.1 as the studied LLM.

Related experimental results are visualized in Figure \ref{fig:pos ratio_diversevul} and Figure \ref{fig:pos ratio_draper}.
We find when the positive sample ratio across the dataset is no less than 30\%,  there is an evident performance gain on F1-Score metric and the recall metric.
Note that higher recall rate indicates that there are less vulnerable samples are predicted to be benign.
This reflects the obvious sensitivity to class imbalance exists in medium-size sequence models like CodeBERT and LLMs like Llama-3.1.
\textit{Therefore, for future studies, we suggest conducting experiments on more balanced datasets, which can help these models to achieve more satisfying performance.}

\begin{figure*}[htbp]
\begin{center}
\includegraphics[width=1.0\textwidth]{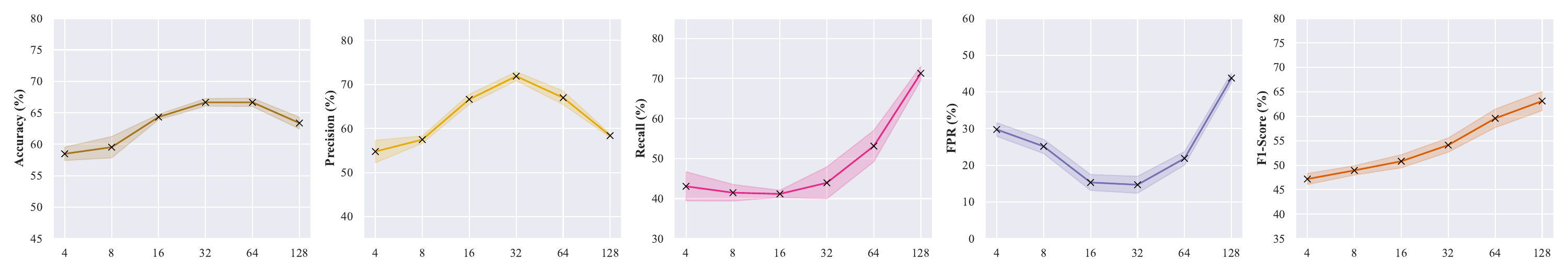}
\caption{Sensitivity Study on LoRA Rank}
\label{fig:rank}
\end{center}
\end{figure*}

\begin{figure*}[htbp]
\begin{center}
\includegraphics[width=1.0\textwidth]{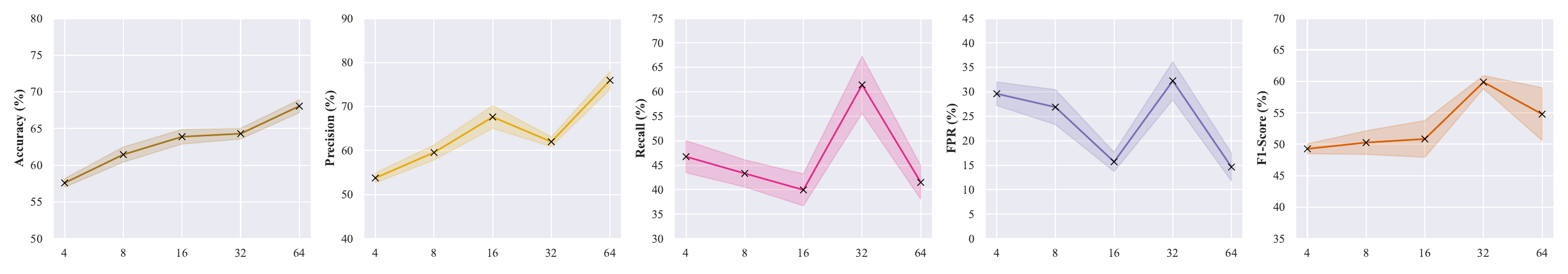}
\caption{Sensitivity Study on LoRA Scaling Factor}
\label{fig:scaling factor}
\end{center}
\end{figure*}

\subsection{Analysis on Analysis on Datasets with Varying Sample Lengths}
\label{sec:length}

In the main experiment, the LLM performed very differently on the long and short sample parts of the same dataset. Because the long and short sample parts in the main experiment have different positive sample ratios, we cannot determine definitively whether sample length or positive sample ratio is the 
 significant factor of influencing model's performance. Thus, we conducted this experiment to investigate the effect of sample length in fine-tuning LLMs.

We divide the sample lengths into 8 intervals, ranging from 0 to 1024, with a step size of 128, ensuring that the number of samples in each length interval is equal. Due to the lack of long samples, we could not subsample for each length interval from a single dataset. As a result, we mix all the 5 datasets. After mixing, we subsample on the mixed dataset to create 8 subsets for every length interval, each with 10,000 samples and 20\% positive sample ratio. The experiments are conducted using Llama-3.1, and the results are shown in Figure \ref{fig:length}. 

In general, sample length has some influence on fine-tuning LLMs. We discover that as sample length increases, the F1-score of the fine-tuned LLM decreases, though this trend is less pronounced than the effect of positive sample ratio discussed in Section \ref{sec:imbalance}. \textit{It can be concluded that positive sample ratio has a much greater impact on fine-tuning the LLM than sample length.}

\subsection{Sensitivity Study}
Our LLM fine-tuning method Low-Rank Adaptation (LoRA) has two important hyper-parameters, i.e. LoRA rank and scaling factor \cite{lora}. 
LoRA rank determines the dimensional characteristics of the matrix after low-rank decomposition, which balances the information capacity, fitting ability and computational cost when fine-tuning the model. 
The scaling factor in LoRA is used to control the magnitude of the low-rank adaptation part of the original pre-training model weight update, thereby balancing the contribution between pre-training knowledge and new task adaptation, and helping the model to adapt downstream tasks more efficiently during fine-tuning. 
We use the F1-score as the main metric across the analysis while other metrics also assist to understand the performance gains. Llama-3.1 is selected as the studied model in the following experiments.

\textbf{Analysis on LoRA Rank.}
We conduct 6 sets of experiments with the rank incrementally set to 4, 8, 16, 32, 64, and 128. The scaling factor is kept equal to the rank in each experiment, following the practice of \cite{lora}. The results are shown in Figure \ref{fig:rank}. 
As we increase the LoRA rank, we find the F1-Score also increases. Therefore, for future studies, a larger LoRA rank is suggested if the computation resource is enough, since a larger rank costs more GPU virtual memory during the fine-tuning process.

\textbf{Analysis on LoRA Scaling Factor.}
We conduct 5 sets of experiments with the scaling factor incrementally set to 4, 8, 16, 32 and 64. The rank is fixed at 16. The results are shown in Figure \ref{fig:scaling factor}. 
As the scaling factor increases, the F1-score first rises and then decreases, peaking at 32. Thus, we reckon a moderate scaling factor is enough during the fine-tuning process, and setting the scaling factor to twice the rank typically yields the best results.

\section{Discussion \& Conclusion}
\label{sec:conclusion}
In this work, we conduct a comprehensive benchmark study towards the code vulnerability detection (CVD) task. We implement 3 graph-based models, 2 medium-size sequence models and 4 open-sourced large language models (LLMs). 
We systemically evaluate the model performance on the long code samples, which are less studied in previous works. 
We identify the class imbalance is a key factor which hinders the performance of LLMs and other models with quantitative experiments, and the sample length of CVD datasets also has a certain impact on fine-tuning LLMs. The sensitivity of 2 main hyperparameters of LoRA\cite{lora} are analyzed in our work. We provide all related codes and resources to facilitate related communities.

For limitations of this work, we do not incorporate the specific prompting techniques like chain-of-thought and in-context learning which some existing literature \cite{xinzhou,vulllm,nong} already focus on. For evaluation on close-source LLMs, we find one helpful Github repository\footnote{\href{https://github.com/soarsmu/ChatGPT-VulDetection}{https://github.com/soarsmu/ChatGPT-VulDetection}} provided in \cite{xinzhou}. 
Furthermore, we don't investigate other parameter-efficient fine-tuning (PEFT) methods, such as QLoRA \cite{qlora}, or full fine-tuning methods.

For future works, as our analysis indicates, class imbalance is one of key factors for this task. The label noise issue also matters as discussed in Section \ref{sec:datasets_detail}.
We aim to investigate the data quality assessment and robust training techniques tailored for the CVD task referring to \cite{dataQuality,dataInconsistent,spw,fnbench,fedelc,fedlsr}, and evaluate the performance of more LLMs with larger parameter space and different architectures (e.g. Deepseek series \cite{deepseek} and Mistral series \cite{mistral}) to study scaling laws.
Furthermore,  we aim to enhance the detection performance of LLMs with the assistance of informative clues \cite{vulllm, unixcoder}, pre-training technique \cite{stagedvulbert} and more continuously updating high-quality dataset \cite{megavul} or more balanced data generation and training techniques \cite{fd,fedlf,fedcrac,smote}. 
If there are some available high-quality and well-curated data, we reckon other effective post-training techniques like direct preference optimization (DPO) \cite{dpo} are expected to further enhance the LLMs' detection precision to identify vulnerable codes, which calls for more joint efforts in future.



\section{Acknowledgemnets}
This work is supported by the National Key Research and Development Program of China (2021YFB2900102) and the National Natural Science Foundation of China (No. 62472410).




\begin{IEEEbiography}[{\includegraphics[width=1in,height=1.25in,clip,keepaspectratio]{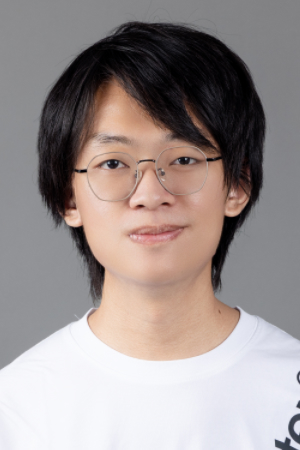}}]{Xuefeng Jiang}
is currently a Ph.D. candidate
with the Institute of Computing Technology, Chinese Academy of Sciences. Before that, he received his bachelor's degree with honors in Beijing
University of Posts and Telecommunications. His
research interests include distributed optimization and machine learning.
\end{IEEEbiography}

\begin{IEEEbiography}[{\includegraphics[width=1in,height=1.25in,clip,keepaspectratio]{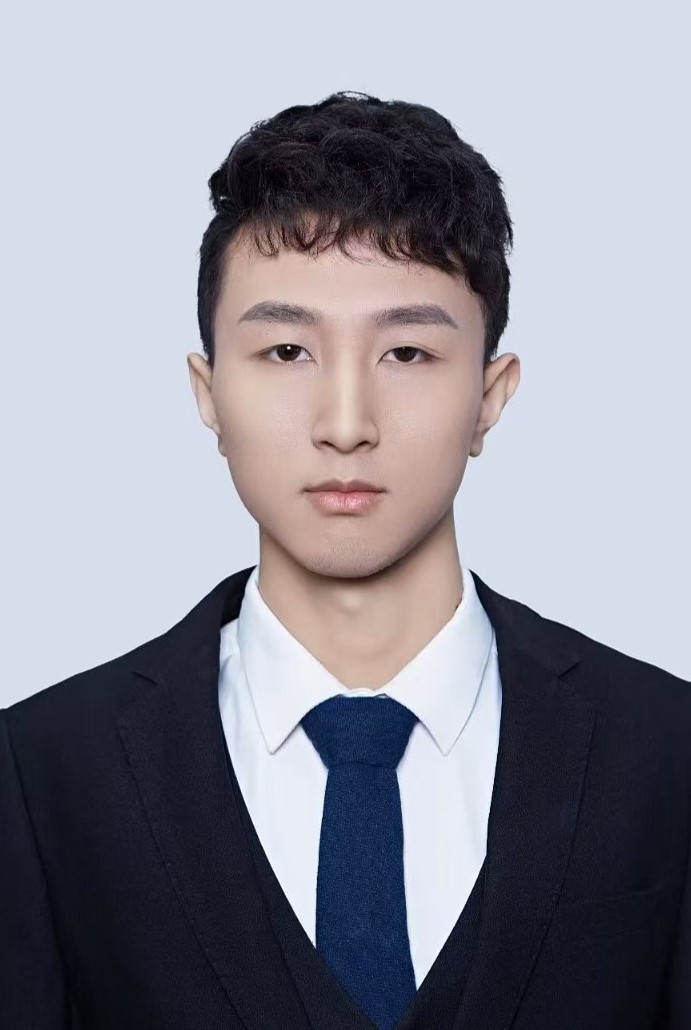}}]{Lvhua Wu}
is currently a master student
with the Institute of Computing Technology, Chinese Academy of Sciences. Before that, he received his bachelor's degree in Beijing
University of Posts and Telecommunications. His
research interests include computer vision, cyber security and machine learning.
\end{IEEEbiography}

\begin{IEEEbiography}[{\includegraphics[width=1in,height=1.25in,clip,keepaspectratio]{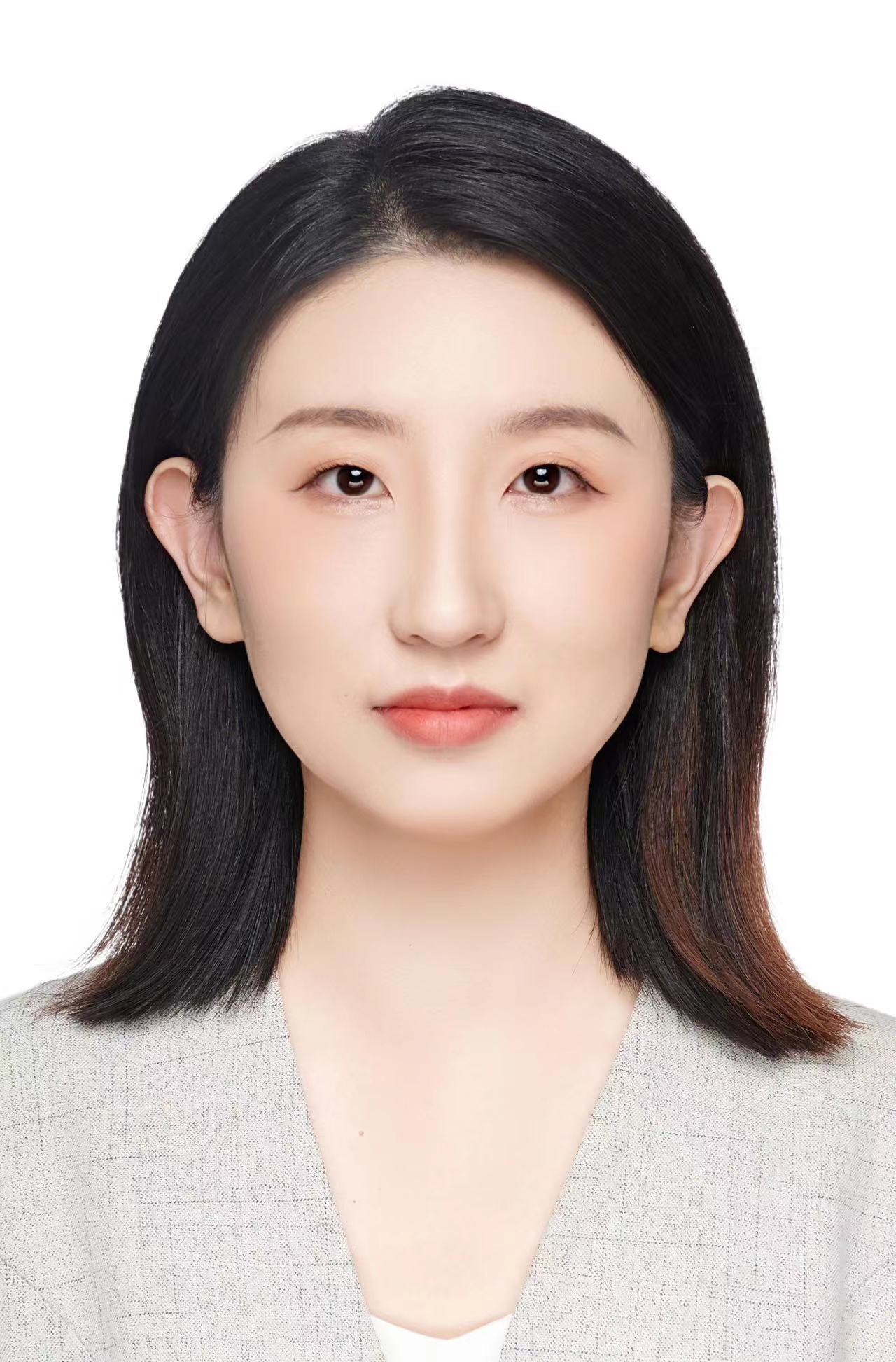}}]{Sheng Sun}
 received her B.S. and Ph.D. degrees
in computer science from Beihang University,
China, and the University of Chinese Academy
of Sciences, China, respectively. She is currently
an associate professor at the Institute of Computing Technology, Chinese Academy of Sciences,
Beijing, China. Her current research interests include federated learning, mobile computing and
edge intelligence
\end{IEEEbiography}

\begin{IEEEbiography}[{\includegraphics[width=1in,height=1.25in,clip,keepaspectratio]{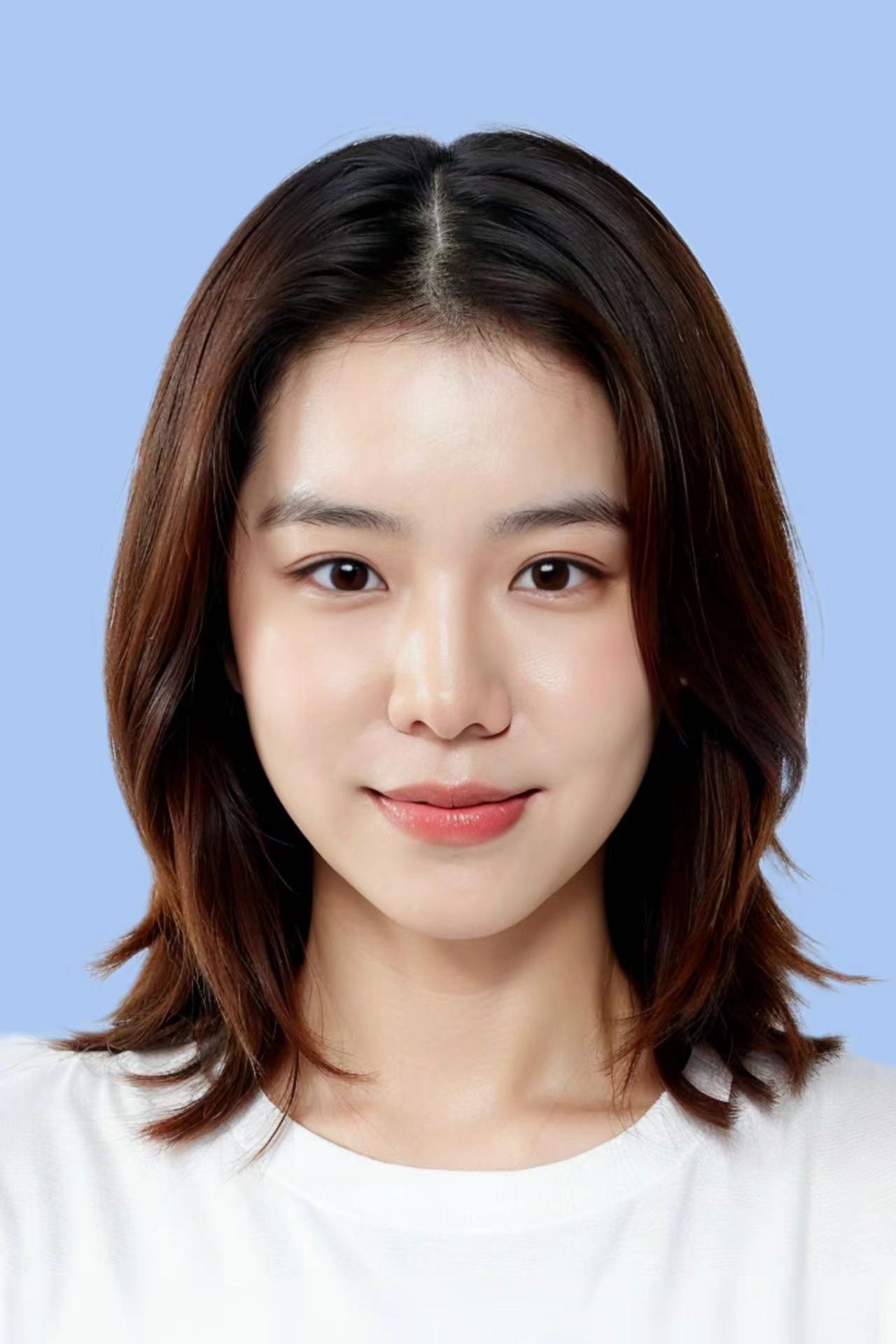}}]{Jia Li}
is currently an algorithm engineer in JD.com, Inc. Before that, she graduated as a master student at Institute of Information Engineering, University of Chinese Academy of Sciences. Her research interests include causal inference, trustworthy AI and machine learning.
\end{IEEEbiography}

\begin{IEEEbiography}[{\includegraphics[width=1in,height=1.25in,clip,keepaspectratio]{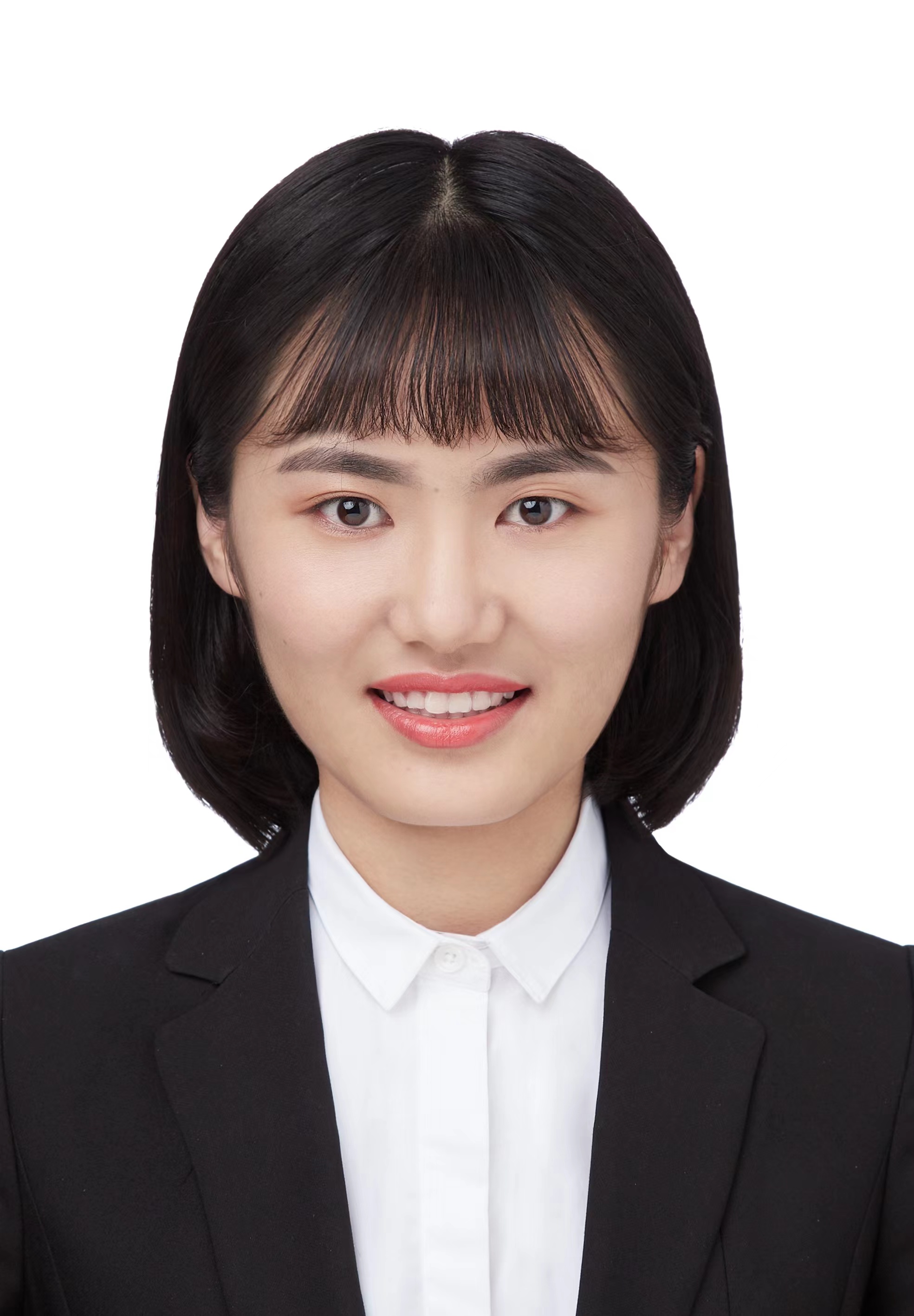}}]{Jingjing Xue} received her B.S. degree from the School of Computer \& Communication Engineering, University of Science and Technology Beijing, China, in 2020. Since 2020, She is currently a Ph.D candidate at the Networking Technology Research Centre, Institute of Computing Technology, Chinese Academy of Sciences. Her current research interests include federated learning and edge intelligence.
\end{IEEEbiography}

\begin{IEEEbiography}[{\includegraphics[width=1in,height=1.25in,clip,keepaspectratio]{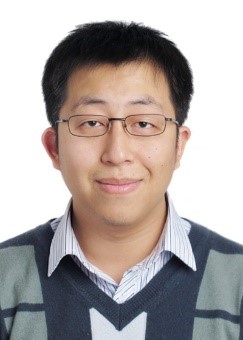}}]{Yuwei Wang}(Member, IEEE) received his Ph.D. degree in computer science from the University of Chinese Academy of Sciences, Beijing, China. He is currently an associate professor at the Institute of Computing Technology, Chinese Academy of Sciences. He has been responsible for setting over 30 international and national standards, and also holds various positions in both international and national industrial standards development organizations (SDOs) as well as local research institutions, including the associate rapporteur at the ITU-T SG21 Q5, and the deputy director of China Communications Standards Association (CCSA) TC1 WG1. His current research interests include federated learning, mobile edge computing, and next-generation network architecture.
\end{IEEEbiography}

\begin{IEEEbiography}[{\includegraphics[width=1in,height=1.25in,clip,keepaspectratio]{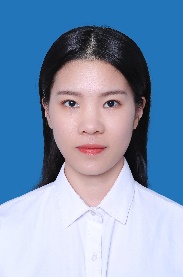}}]{Tingting Wu}
 received the Ph.D. degree from the Shenyang Institute of Automation Chinese Academy of Sciences in 2023. She is currently a senior researcher at the China Mobile Research Institute, her main research interests include neural network compression, distributed intelligence, data privacy protection, large model privacy protection, etc., and has published more than ten journal articles and conference papers.
\end{IEEEbiography}

\begin{IEEEbiography}[{\includegraphics[width=1in,height=1.25in,clip,keepaspectratio]{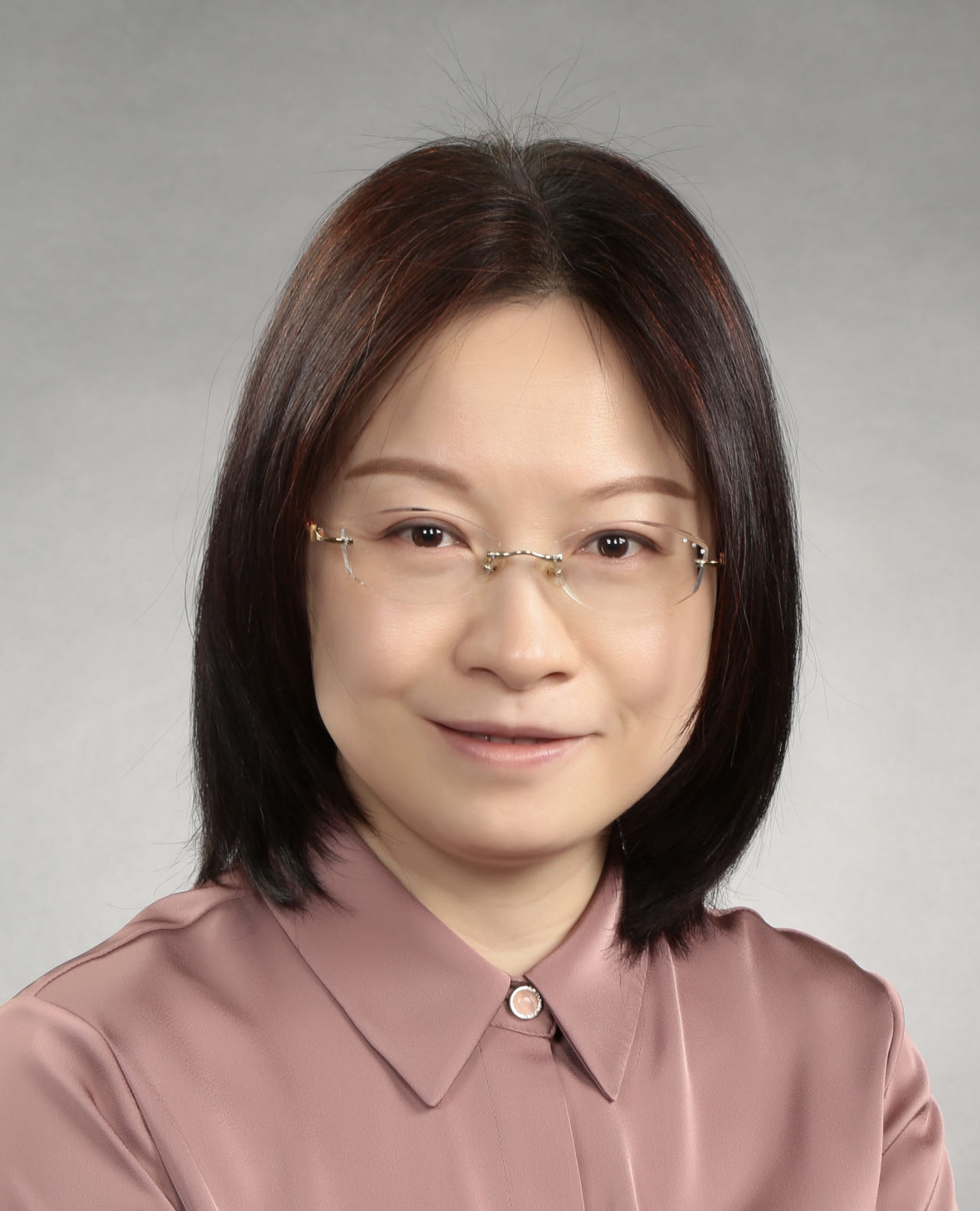}}]{Min Liu} 
(Senior Member, IEEE) received her Ph.D degree in computer science from the Graduate University of the Chinese Academy of Sciences, China. Before that, she received her B.S. and M.S. degrees in computer science from Xi’an Jiaotong University, China. She is currently a professor at the Institute of Computing Technology, Chinese Academy of Sciences, and also holds a position at the Zhongguancun Laboratory. Her current research interests include mobile computing and edge intelligence.
\end{IEEEbiography}

 





\end{document}